\documentclass{hld2025}

\title[Learning Dynamics of Multi-head Latent Attention]{A Random Matrix Theory Perspective on the Learning Dynamics of Multi-head Latent Attention}

\usepackage{graphicx}
\usepackage{pifont}
\usepackage{enumitem}
\usepackage{listings}
\usepackage{booktabs}
\usepackage{array}
\usepackage{wrapfig}

\hldauthor{%
\Name{Nandan Kumar Jha} \Email{nj2049@nyu.edu}\\
\addr {New York University} 
\AND
\Name{Brandon Reagen} \Email{bjr5@nyu.edu}\\
\addr {New York University} }

\begin{document}

\maketitle

\begin{abstract}
In this work, we study how multi-head latent attention (MLA), a popular strategy for compressing key/value memory, affects a transformer's internal capacity during pretraining. Using a lightweight suite of Marchenko–Pastur (MP) diagnostics, we analyze the spectrum of the $W_{Q}W_{K}^\top$ gram matrix throughout training, comparing three variants: the standard multi-head attention (MHA) baseline, MLA-PreRoPE with rotary applied before compression, and MLA-Decoupled, which shares a single rotary sub-vector across all heads. Our random matrix analysis reveals {\bf three key findings:} {\bf i)} capacity bottlenecks emerge locally: both MHA and MLA-PreRoPE exhibit sharp, early spikes in specific layers that persist and propagate, disrupting the balance between bulk and outlier directions; {\bf ii)} these spikes coincide with rank collapse, concentrating the model's expressivity into narrow subspaces; {\bf iii)} only the decoupled variant prevents this cascade, maintaining broad spectral support and suppressing outlier formation across layers. These results underscore that \emph{how} rotary embeddings are applied is just as critical as \emph{where} compression occurs. Sharing rotary components across heads mitigates spectral fragmentation and preserves representational capacity. 
\end{abstract}

\section{Introduction}

Modern large language models (LLMs) continue to grow in scale and capability; however, their practical utility is constraint by increasing inference latency stems from memory-bound key/value (KV) cache operations, rather than compute. To address this, recent architectures such as DeepSeek-V2/V3 have introduced Multi-head Latent Attention (MLA) \cite{liu2024deepseek,liu2024deepseekV3,zhao2025insights,meng2025transmla}, which compresses queries and keys into lower-dimensional latent representations before attention computation. This design reduces KV cache size by over 50\% while maintaining strong performance.

Despite these practical gains, the fundamental question remains unanswered: {\em How does latent compression impacts the spectral dynamics of attention, and what are its implications for learning and generalization?} While Random Matrix Theory (RMT) has emerged as a powerful tool to study neural network internal dynamics \cite{dandi2025a,firdoussi2025maximizing,thamm2024random,bouchard2024random,ilbert2024analysing,levi2023underlying,adlam2022random,martin2021implicit,staats2024locating,feofanov2023random,wei2022more,tiomoko2019random,pennington2017nonlinear,liao2018dynamics,couillet2016random,pennington2017geometry}, its application to the spectral behavior of attention mechanisms under latent-space compression remains largely unexplored. 

This gap limits our understanding of MLA's inductive biases and potential pitfalls. For instance, do spectral pathologies such as rank collapse persist under MLA? Can width compression alone prevent outlier growth, or do we need additional design choices, such as applying rotary embeddings before compression? While prior studies have quantified the memory efficiency of MLA, they often overlook the underlying spectral dynamics that contribute to its efficiency. 

In this work, we aim to bridge this gap by investigating following research questions: \\
{\bf RQ1:} Where do MLA-induced spectral spikes emerge? Are they layer- or head-specific? \\ 
{\bf RQ2:} {Is latent compression alone sufficient to suppress outliers, or rotary-vector sharing matters?} \\ 
{\bf RQ3:} {What impact do residual spikes have on rank collapse and latent space utilization?}

We summarize our {\bf contributions} as follows. 
\begin{enumerate} [noitemsep,nolistsep,leftmargin=0.5cm] 
  \item {\em RMT-based diagnostic framework for attention.}  We develop a lightweight tool to analyze the squared singular values spectrum of  $W_Q W_K^\top$ Gram matrix, using four Marchenko-Pastur (MP) metrics \citep{marchenko1967distribution}: MP-Gap, outlier count and energy, MPSoft rank, and stable rank.
     
  \item {\em First spectral analysis of MLA during training.}  We apply our framework to benchmark classical MHA and two MLA variants: MLA-PreRoPE, where rotary embeddings are applied before up-projection, and MLA-Decoupled, which uses a shared rotary vector across heads. For consistency and fair evaluation, all experiments are conducted within the LLaMA architecture.

  \item {\em Identification of a mid-layer spike cascade.}  
  We discover that spectral spikes emerge early and persist in MHA and MLA-PreRoPE models, causing severe rank collapse. PreRoPE partially mitigates but does not eliminate these effects.

  \item {\em Rotary-vector sharing as a key mitigation strategy.}  
      The MLA-Decoupled variant suppresses spectral outliers: maintains a low MP-Gap, and preserves stable rank across layers. This highlights that rotary-vector sharing is crucial for maintaining a good spectral behavior.
\end{enumerate}

\section{Experimental Setup}

To investigate how latent compression and rotary embedding strategies influence the spectral dynamics of the attention mechanism, we perform  RMT analytics on $W_Q W_K^\top$ Gram matrix \cite{bao2024self} at each attention layer, using four MP metrics: MP-gap, outlier count and energy, and soft and stable rank.

We integrate all three variants into a LLaMA-130M architecture and train them from scratch for 20K steps on 2.2B tokens from the C4 dataset, using a context length of 256. We adopt the  downscaled architectural settings and training setup for LLaMA-130M from \cite{li2025mixln}. Training is performed with a global batch size of 512 on two RTX 3090 GPUs (24 GB each). In both MLA variants, we apply a compression ratio of 2, reducing the latent dimension from 64 to 32. 

\section{Spectral Analysis of Latent Compression in MLA via Random Matrix Theory}

{\bf Notations}  $W_Q$ and $W_K$ denote the query and key weight matrices, respectively; $d_{\text{model}}$ is the model embedding dimension; $H$ is the number of attention heads; and $d_k$ is the per-head dimension.

\paragraph{Cross-Gram construction}
For each attention layer we consider the learned query and key projection weights $W_Q, W_K \in \mathbb{R}^{m \times d_{\text{in}}}$, where $m = H \cdot d_k$ is the total head dimension and $d_{\text{in}}$ is the input dimension of the projection matrices. At every logging step we form the \emph{cross-Gram} matrix 
\vspace{-0.5em}
\begin{equation} 
G = \frac{1}{d_{\text{in}}} W_Q W_K^{\top} \in \mathbb{R}^{m \times m},
\tag{1}
\end{equation}
and compute its eigenvalues via singular-value decomposition (SVD). 
\vspace{-0.5em}

\paragraph{Rationale}
For MHA, the input dimension to the query/key projection is $d_{\text{in}} = d_{\text{model}}$ (e.g., 768 in LLaMA-130M). In MLA variants, this projection is factorized into a shared down-projection ($W^{\downarrow}$) and a head-specific up-projection $(W_Q^{\uparrow}, W_K^{\uparrow})$. We analyze {\em only the up-projection}, setting $d_{\text{in}} = 32$,  to {\em isolate} the effects of latent compression and rotary embedding design on the attention spectrum, while keeping the shared $W^{\downarrow}$. In the \emph{decoupled} setting, we further isolate the RoPE branch, with $d_{\text{in}} = 32$ and row dimension $m = \frac{1}{2} H d_k$, to focus purely on the query/key structure without confounding from the value pathway.

\paragraph{Marchenko--Pastur (MP) metrics}
Dividing by $d_{\text{in}}$ sets the expected entry variance of $G$ to one under the i.i.d.\ null model. With aspect ratio $\gamma = m / d_{\text{in}}$, the MP bulk edges are therefore
\begin{equation}
\lambda_{\pm} = (1 \pm \sqrt{\gamma})^{2}.
\tag{2}
\end{equation}


\begin{wraptable}[10]{r}{0.51\textwidth - .25\columnsep}
\centering
\vspace{-0.85\intextsep}
\resizebox{0.52\textwidth}{!}{
\begin{tabular}{@{}lll@{}}
\toprule
Metric & Formula & Interpretation \\
\midrule
MP-Gap & $\Delta = \lambda_1 - \lambda_+$ & Spike strength \\
\addlinespace
Outlier Count & $\#\{\lambda_i > \lambda_+\}$ & Spike population \\
\addlinespace
Outlier Energy & $\frac{\sum_{\lambda_i > \lambda_+} \lambda_i}{\sum_i \lambda_i}$ & Spectral mass lost to spikes \\
\addlinespace
MPSoft Rank \cite{martin2021implicit} & $\rho = \frac{\lambda_1}{\lambda_+}$ & Normalized spike distance \\
\addlinespace
Stable Rank \cite{martin2021implicit} & $r_+ = \sum \frac{\lambda_i}{\lambda_1}$ & Residual capacity \\
\bottomrule
\end{tabular}}\\
\vspace{-0.5em}
\caption{Summary of spectral measures and their interpretations ($\lambda_1$ is largest eigen value)}
\label{tab:spectral_measures}
\end{wraptable}
Table \ref{tab:spectral_measures} summarize the MP metrics used for spectral analysis.  
{\tt MP-Gap} quantifies the strength of the dominant spike: a value of zero indicates that the bulk spectrum lies entirely within the MP bulk, while larger values reflect a detached leading eigenvalue. {\tt Outlier Count} measures how many eigenvalues exceed the MP upper edge, capturing the prevalence of spiking. {\tt Outlier Energy} quantifies the spectral mass of these spikes, translating it into a fractional energy \emph{budget}. Finally, {\tt MPSoft-} and {\tt Stable-Rank} turn spike behavior into capacity metrics: soft-rank  measures the relative distance of the top spike from the bulk edge, while stable-rank  captures how much of the bulk dimension remains after excluding spikes.

\section{Experimental Results}

\begin{figure} [t]
\centering
\subfigure[MP-Gap Comparison  \label{subfig:mp_gap_comparison}]{\includegraphics[width=.32\textwidth]{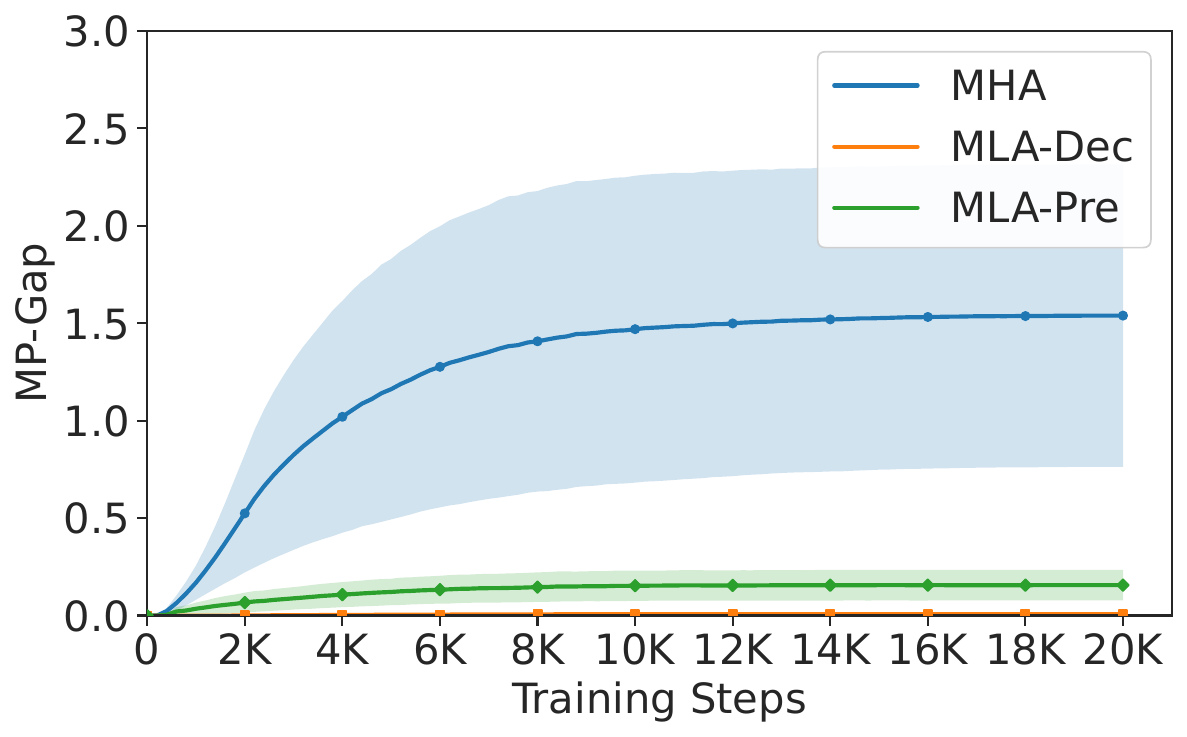}} 
\subfigure[Upper Counts Comparison  \label{subfig:up_cnt_comparison}]{\includegraphics[width=.31\textwidth]{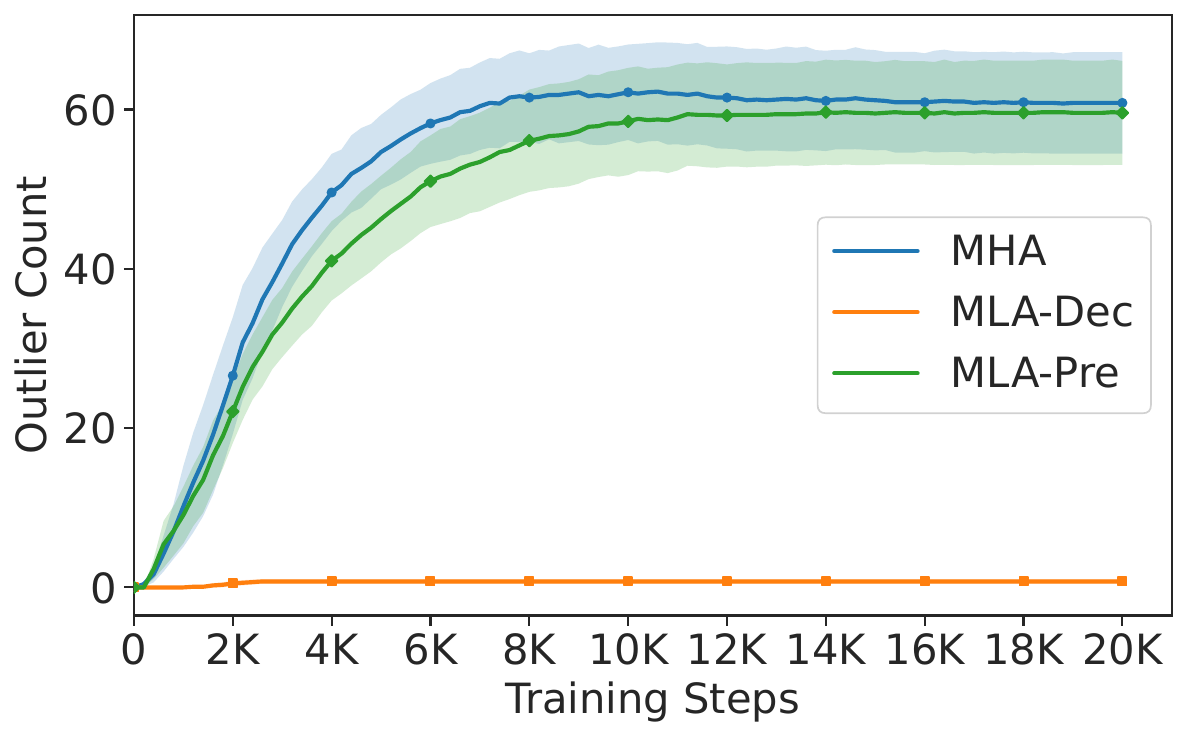}} 
\subfigure[Upper Energy Comparison \label{subfig:up_energy_comparison}]{\includegraphics[width=.31\textwidth]{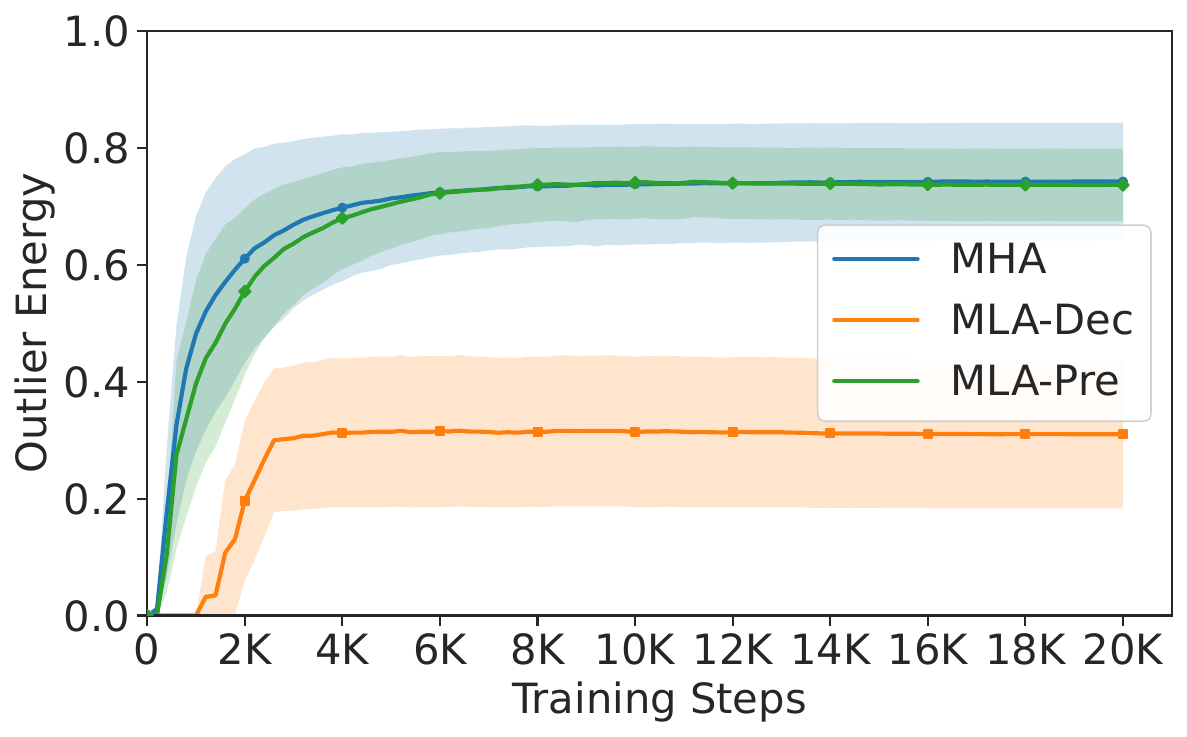}}  \\ \vspace{-1.5em}
\caption{ {\bf Spectral-spike dynamics:} (a) MP-Gap, (b) outlier count, and (c) outlier energy, for MHA (\textcolor{blue}{blue}), MLA-Dec (\textcolor{orange}{orange}), and MLA-Pre (\textcolor{green}{green}). 
Curves show layer-wise means and shaded bands denote $\pm$1  standard deviation in LLaMA-130M.  
Together, these metrics capture the emergence and strength of spectral outliers in the $W_Q W_K^\top$ spectrum.} 
\label{fig:UpperEnergCounts}
\end{figure}

\textbf{Decoupled MLA substantially reduces spectral spikes}
As shown in \ref{fig:UpperEnergCounts}(a), MP-Gap in classical MHA rises to $\approx 2$ within the first 5K steps and then plateaus, indicating a persistent, high-magnitude spectral spike. Pre-RoPE MLA exhibits a similar trend but saturates at  a lower amplitude, suggesting that latent compression alone does not suppress spike formation. In contrast, Decoupled MLA keeps the MP-Gap near-zero throughout training, indicating that its shared rotary sub-vector prevents singular values from escaping the MP bulk.

Figure \ref{fig:UpperEnergCounts}(b)  shows the the number of outlier eigenvalues, exceeding the MP upper edge. MHA and Pre-RoPE both stabilize at 60 to 65 outliers per layer (roughly 5 to 6 per head), while Decoupled MLA consistently exhibits zero, empirically confirming the absence of spectral outliers and validating the collapsed MP-Gap. Figure \ref{fig:UpperEnergCounts}(c)  quantifies outlier energy, the proportion of total spectral energy carried by these spikes. MHA and Pre-RoPE MLA channel nearly 70\% of the spectrum into the spike subspace, signaling severe rank collapse. In contrast, Decoupled MLA re-distributes this energy back into the bulk, dropping outlier energy below 30\%. 

This shift reflects a broader effective rank and reinforces a key insight: {\em how rotary embeddings are applied matters}. Head-shared rotary embeddings suppress the spike formation substantially, whereas conventional key/query compression schemes do not.

\begin{figure} [htbp]
\centering
\subfigure[MP-SoftRank Comparison ]{\includegraphics[width=.4\textwidth]{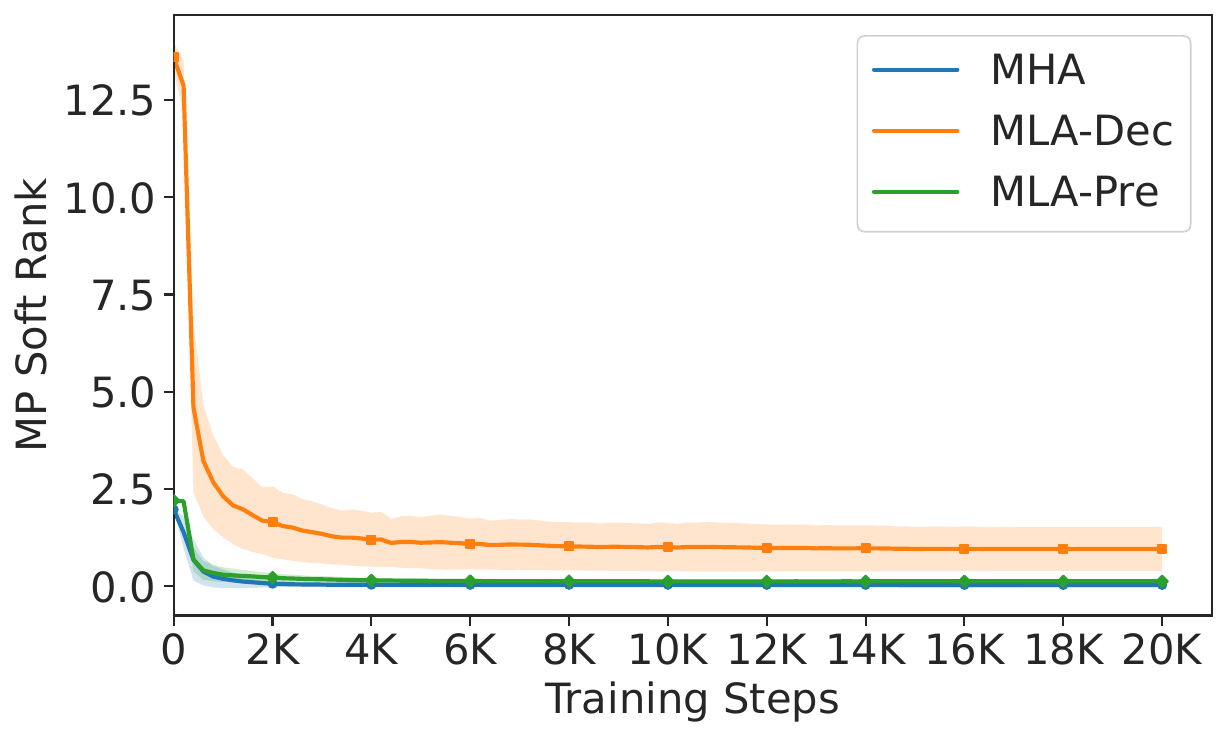}} 
\subfigure[Stable Rank Comparison  ]{\includegraphics[width=.4\textwidth]{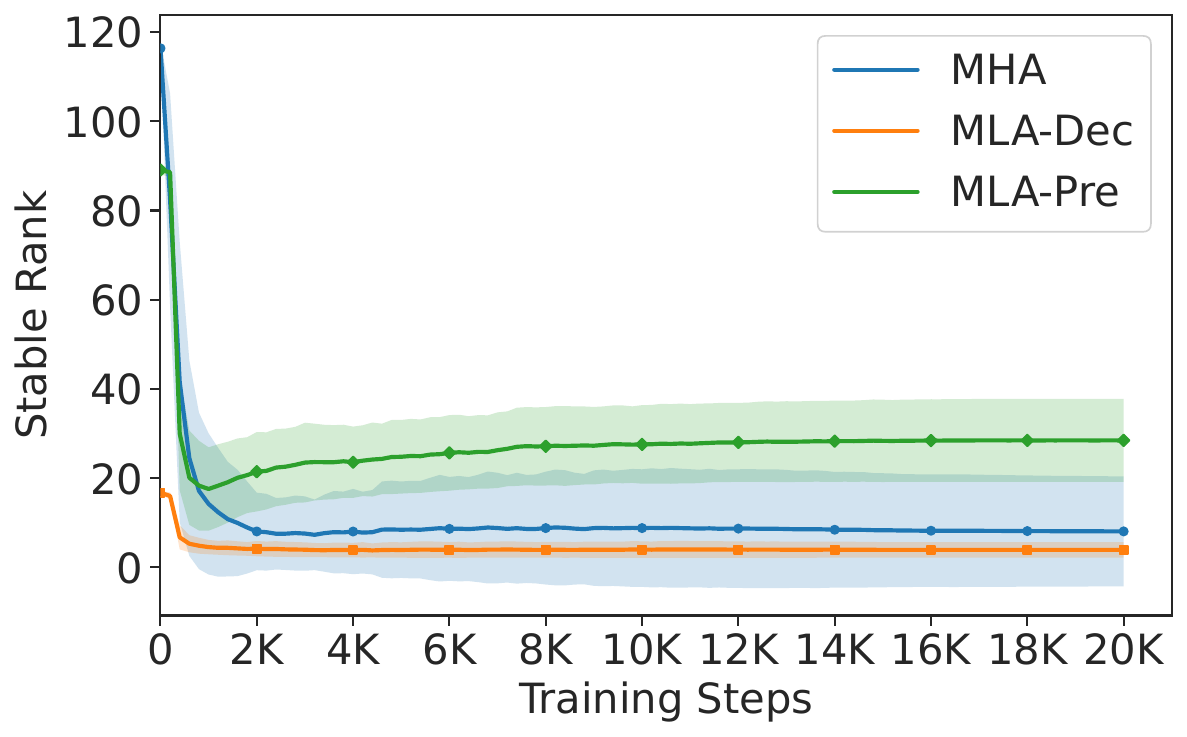}} \vspace{-1em}
\caption{ {\bf Spectral-Capacity Dynamics:} (a) MP-Soft-Rank, and (b) Stable Rank are shown for LLaMA-130M model with MHA (\textcolor{blue}{blue}), MLA-Pre (\textcolor{orange}{orange}), and MLA-Dec (\textcolor{green}{green}). 
Curves show layer means; shaded regions indicate $\pm 1$ standard deviation across 12 layers. 
Higher MP-Soft-Rank signals sharper spectral spikes; higher Stable Rank indicates better bulk direction usage. 
MLA-Dec excels at suppressing outliers, while MLA-Pre offers the highest representational capacity. MHA remains in between on both metrics.
} 
\label{fig:RankLinePlots}
\end{figure}

{\bf Compression-regularization trade-off: Decoupled MLA suppresses outliers while Pre-RoPE MLA maximizes capacity}
Figure \ref{fig:RankLinePlots} illustrates two sides of the spectral trade-off. First, MP-Soft-Rank which measures how far the largest eigenvalue lies above the MP bulk. After just 1K steps, the Decoupled MLA drives this ratio to $\approx 1.0$, substantially reducing spectral spikes (outliers). In contrast, classical MHA and Pre-RoPE MLA stabilize at $\approx 1.2$ and $\approx 1.5$, respectively, indicating persistent spike formation.

Nonetheless,  on the the flip side, the Stable-Rank, a proxy for usable dimensionality shows a reverse order: MLA-Pre retains the highest capacity ($\sim45$), MHA saturates at off around $\sim12$, while MLA-Dec collapses to $\sim5$. This trade-off is expected. The decoupled variant reduces the row dimension (by isolating the RoPE branch), shares a single rotary sub-vector across heads, and applies a strong compression bottleneck ($d_{\text{in}} = 32$). These factors suppress the top singular value---effectively taming spikes---but also reduce the Frobenius norm more rapidly than the spectral norm, leading to lower stable rank. By contrast, MLA-Pre retains the full row dimension and latent energy, preserving many directions even while tolerating a moderate spike.

\begin{figure} [t]
\centering
\subfigure[{\footnotesize MPGap in MHA} ]{\includegraphics[width=.32\textwidth]{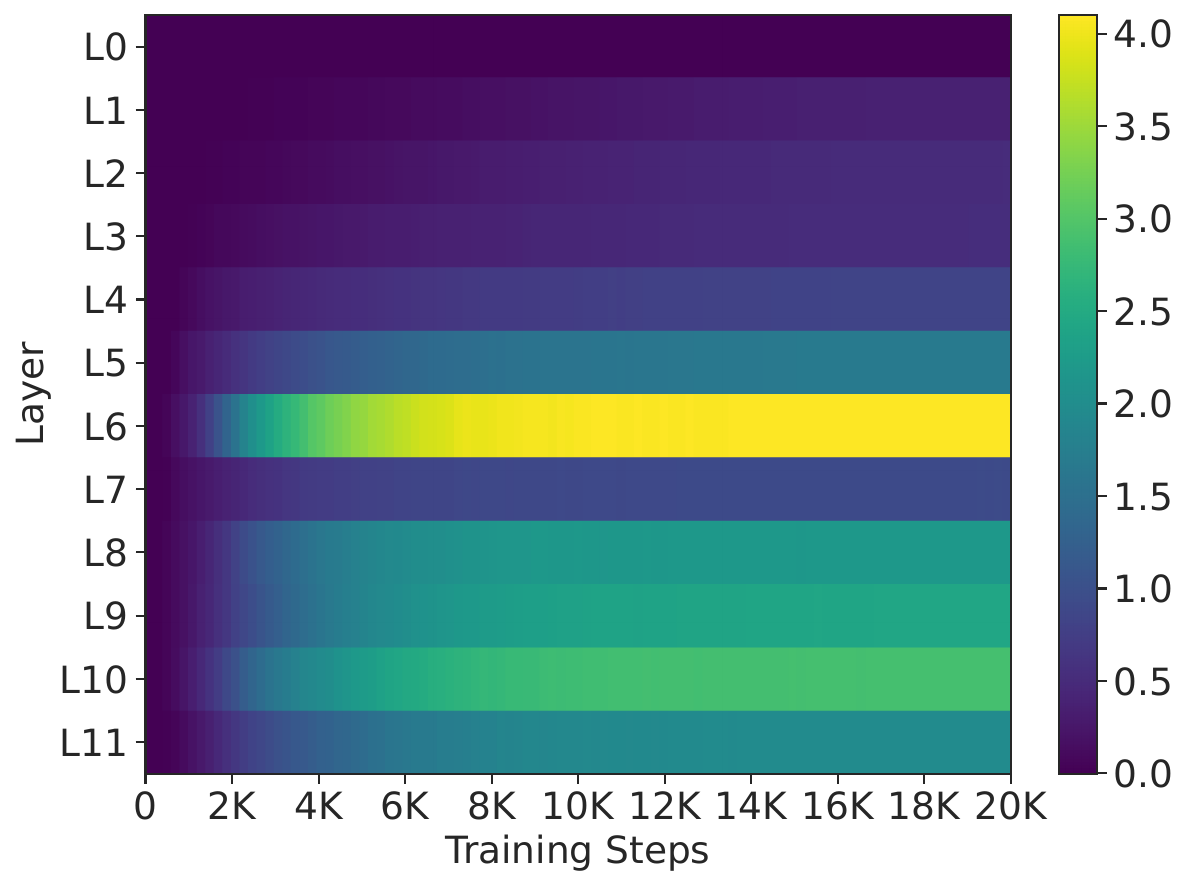}} 
\subfigure[{\footnotesize MPGap in MLA-Decoupled }]{\includegraphics[width=.32\textwidth]{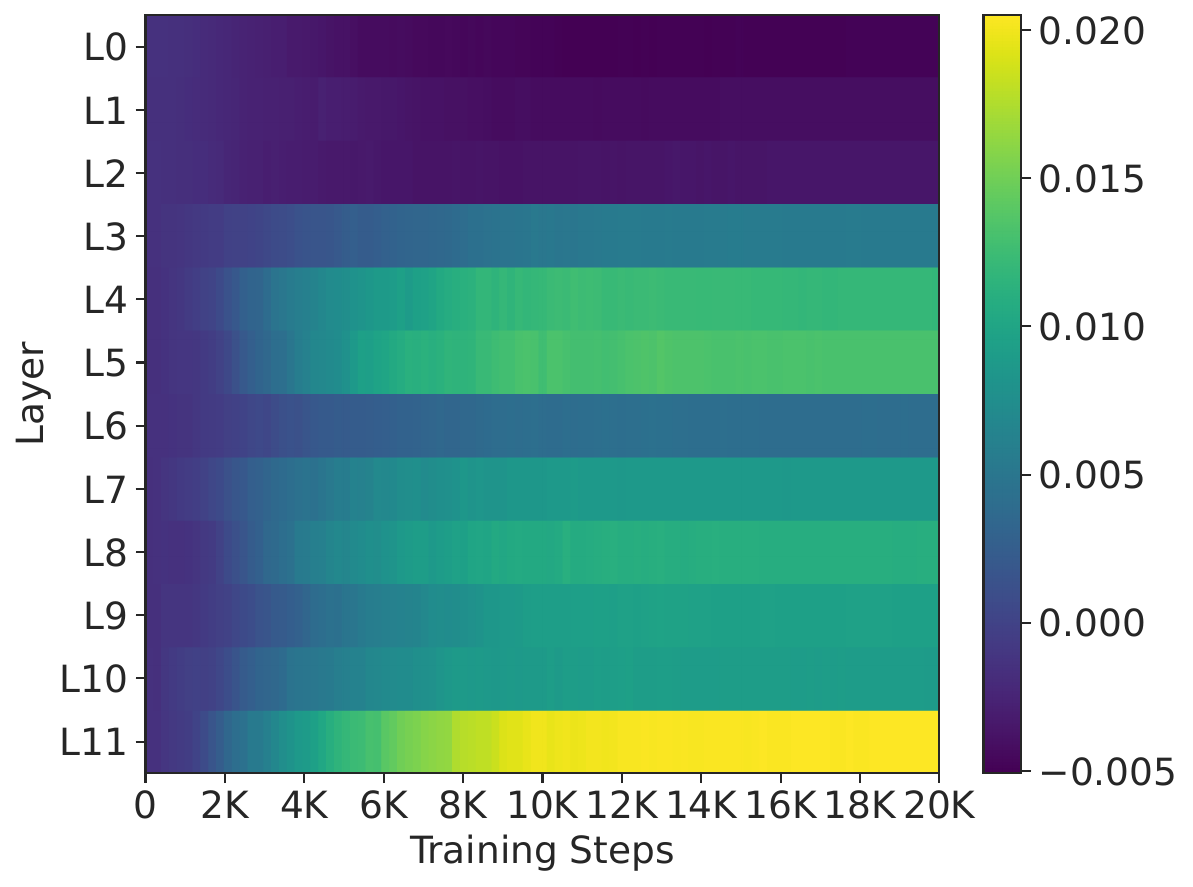}} 
\subfigure[{\footnotesize MPGap in MLA-PreRoPE }]{\includegraphics[width=.32\textwidth]{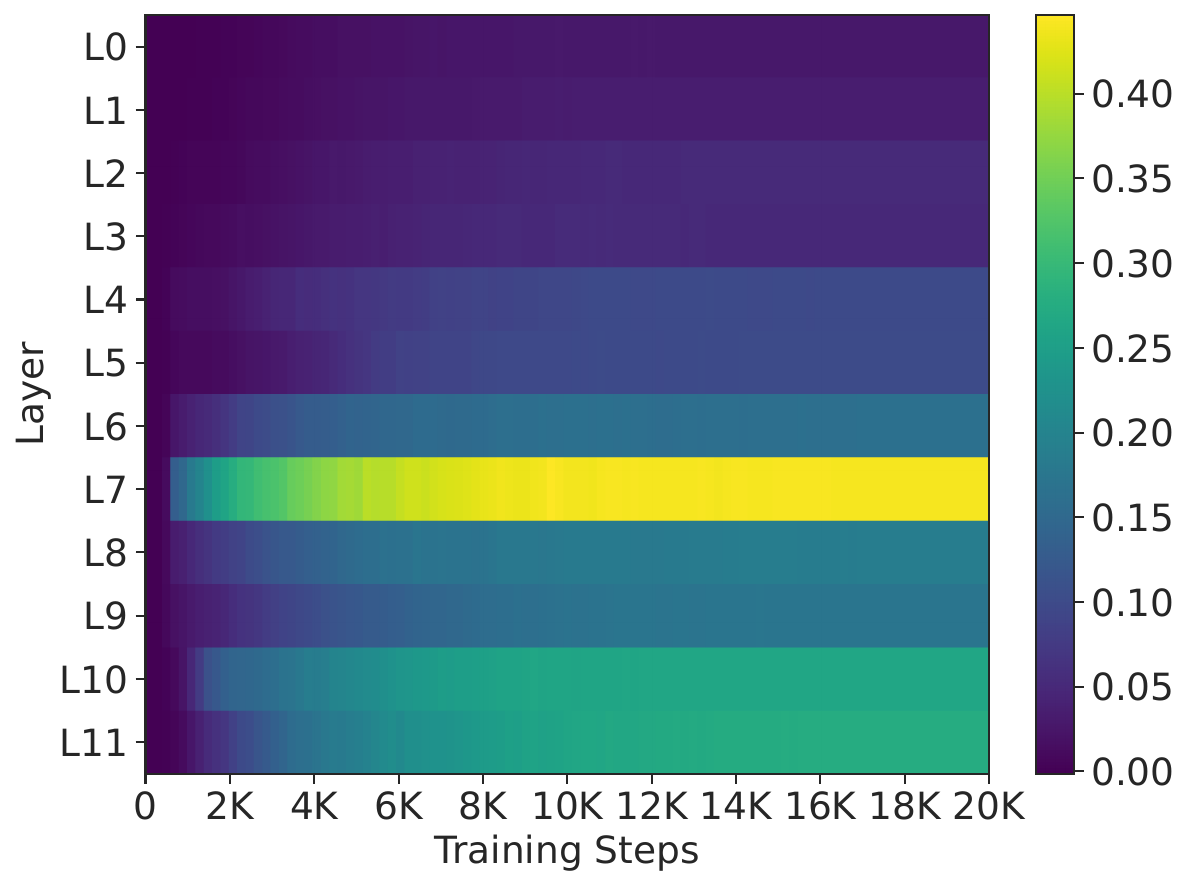}}  \\ 
\subfigure[{\footnotesize StableRank  in MHA }]{\includegraphics[width=.32\textwidth]{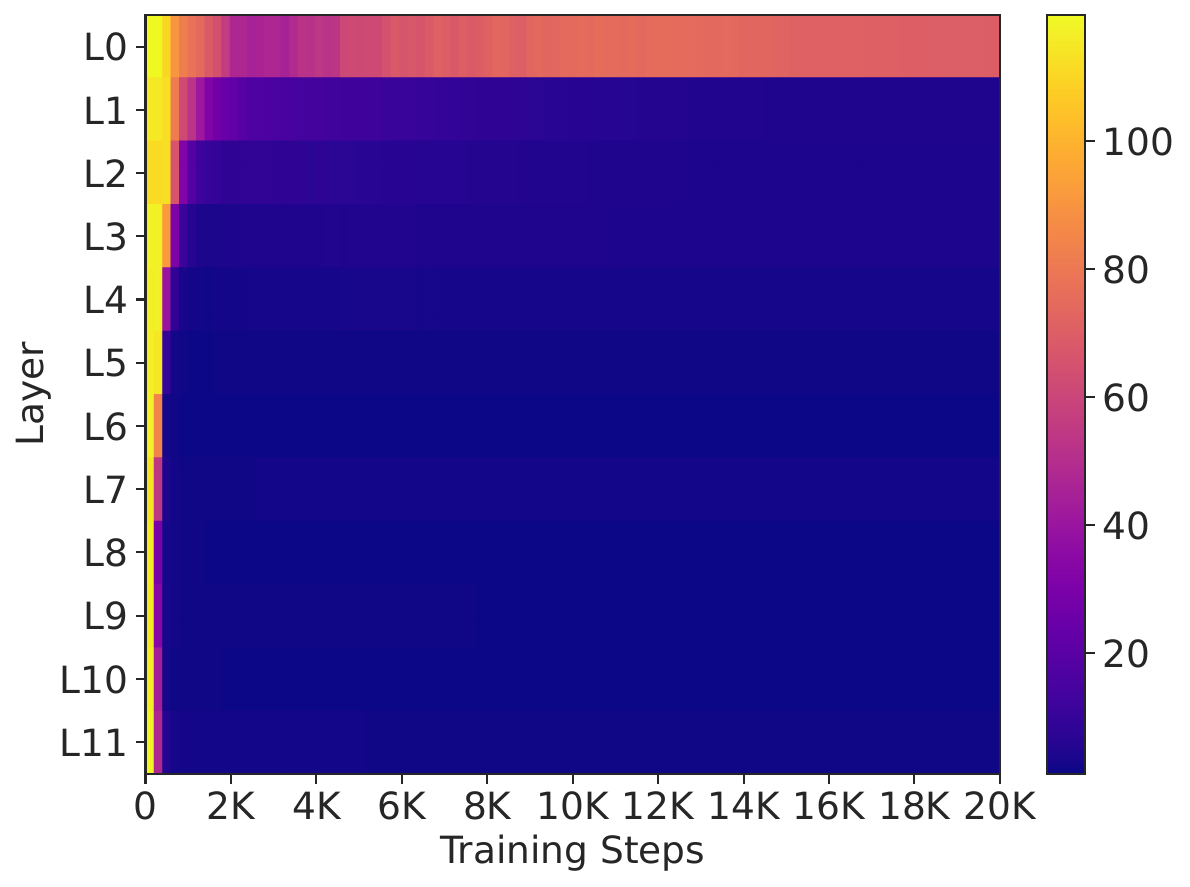}} 
\subfigure[{\footnotesize StableRank in MLA-Decoupled }]{\includegraphics[width=.32\textwidth]{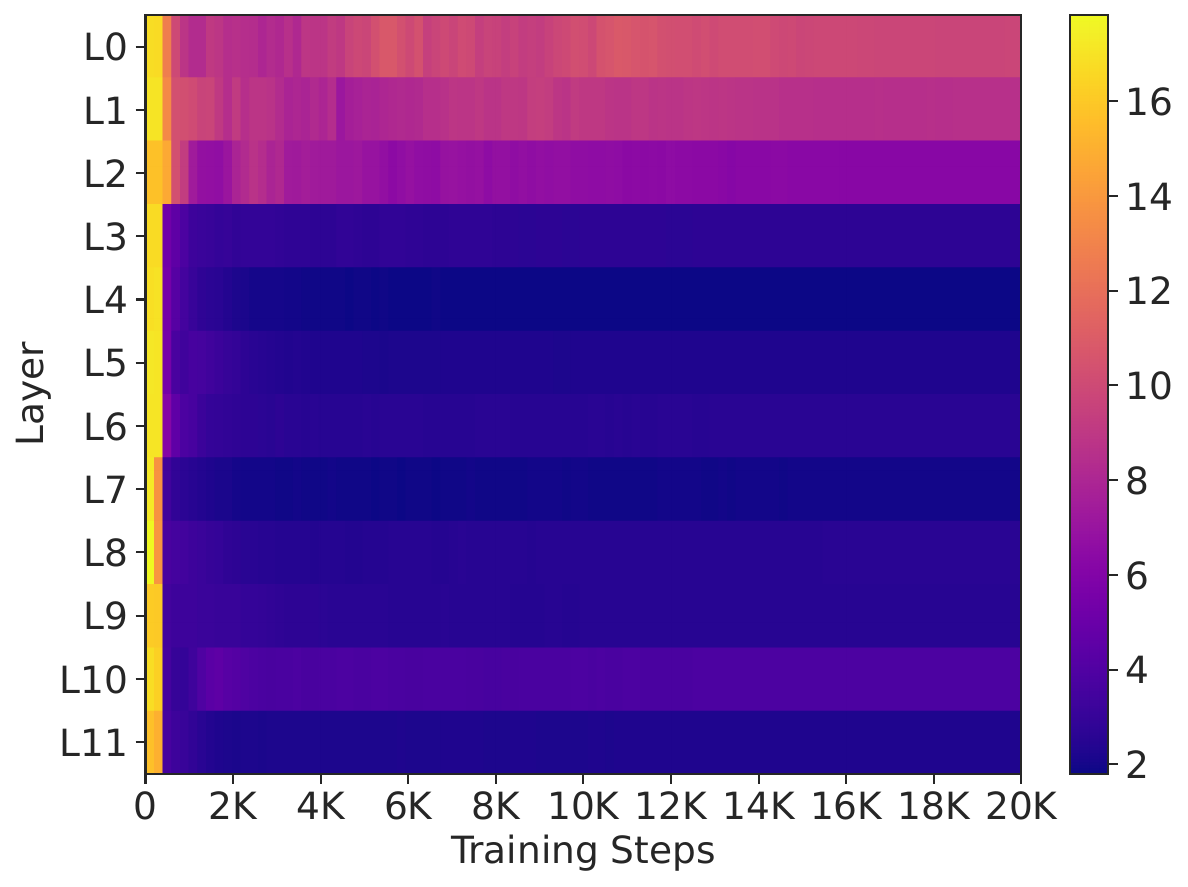}} 
\subfigure[{\footnotesize StableRank in MLA-PreRoPE} ]{\includegraphics[width=.32\textwidth]{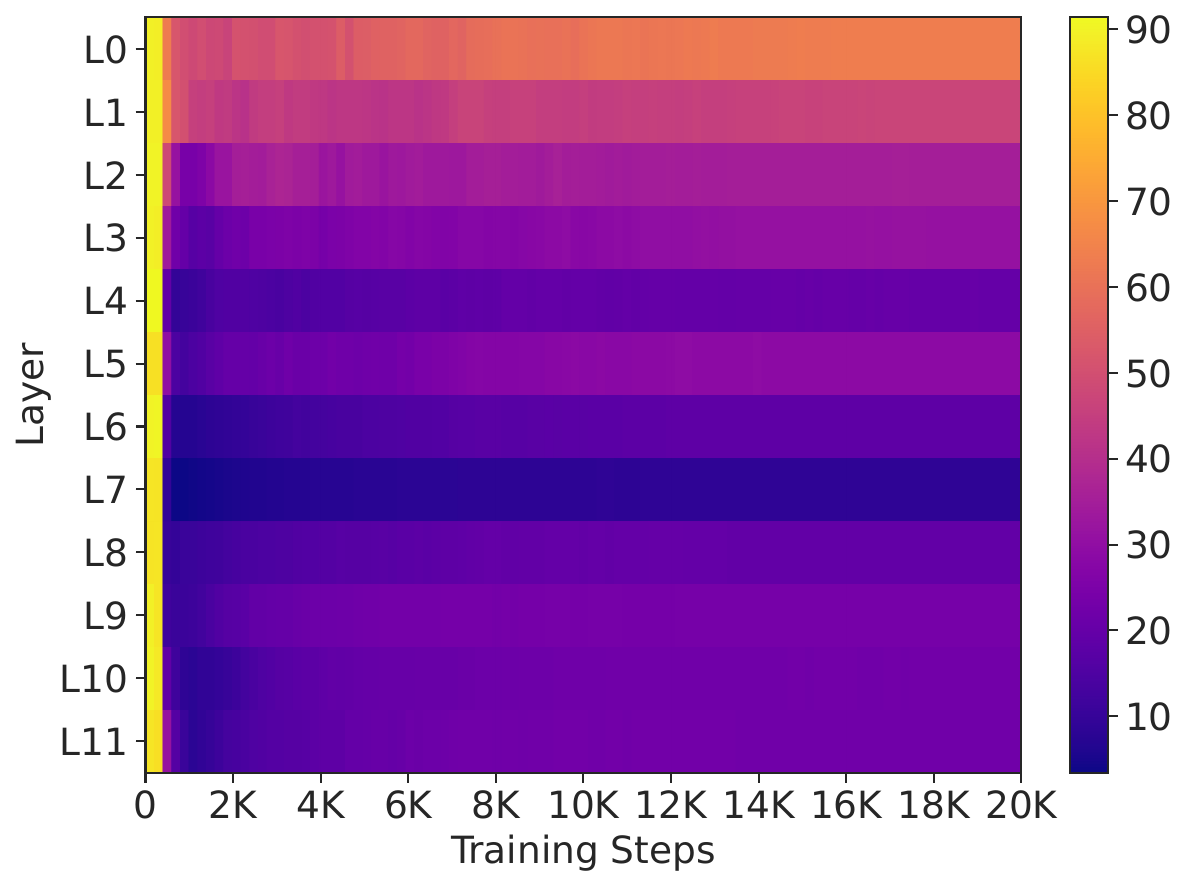}}  \\ \vspace{-1em}
\caption{{\bf Layerwise spectral dynamics}: (Top row)MP-Gap and (bottom row)StableRank heatmaps. MHA exhibits strong mid‐layer concentration in MPGap and declining StableRank in later layers, while MLA‐based methods spread representational changes more evenly, maintaining higher stable ranks across depths } 
\label{fig:MPGapStableRankHeatMaps}
\end{figure}

{\bf Capacity bottlenecks: Mid-layer spikes vs. uniform utilization}
The MP-Gap heatmaps (top row of Figure \ref{fig:MPGapStableRankHeatMaps}) show a sharp {\em hot band} in MHA: the mid layer (L6) reaches to a gap of $\approx4$ within the first 5k steps and the spike then diffuses into deeper layers.  Pre-RoPE MLA shows the same localization but the peak magnitude is roughly one-tenth that of MHA, suggesting that latent compression (by a factor of 2) dampens the spikes  but does not remove them completely.  
By contrast, Decoupled MLA remains essentially flat ($<0.02$ throughout), demonstrating that spreading each rotary sub-vector across heads suppresses edge singular values and prevents spike formation.

The Stable-Rank heat-maps (bottom row) complete the picture.  MHA starts with a high rank ($\sim120$) in the first few layers but collapses below $\sim$20 after layer $\sim$5, mirroring the MP-Gap spike.  Pre-RoPE MLA partially recovers in deeper layers ($\sim$20--30\%) and retains $\sim$40\% rank in earlier layers, though it still underperforms. In contrast, Decoupled MLA consistently sustains $>$60\% normalized rank across all layers and training steps, indicating stable representational capacity.


{\bf Outlier energy distribution shows spectral compression vs. spread}


\begin{wrapfigure}[13]{r}{0.48\textwidth  - .25\columnsep}
\centering
\vspace{-2\intextsep}
\includegraphics[width=.49\textwidth]{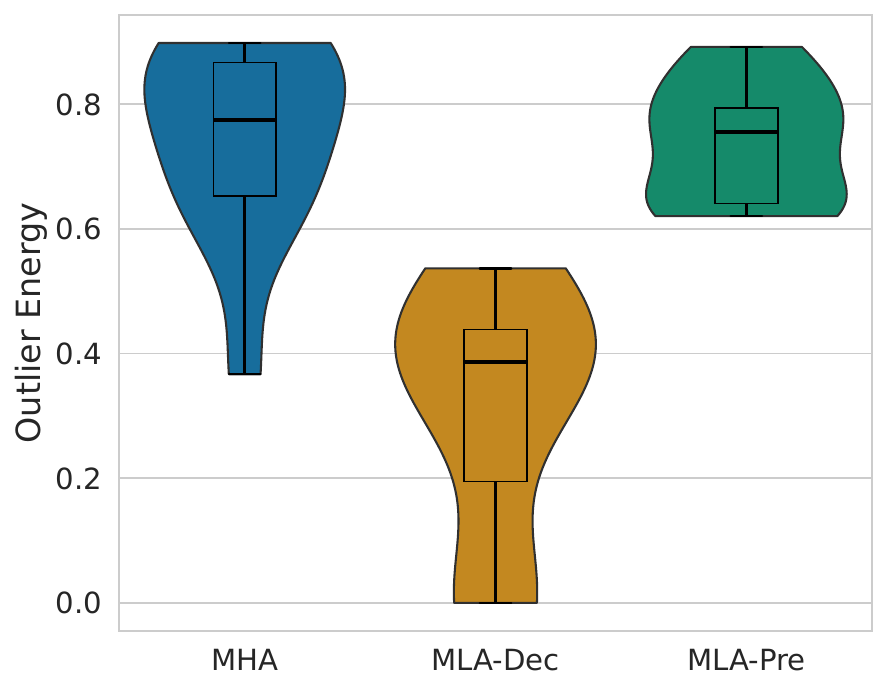} 
\vspace{-2.5em}
\caption{Upper energy  violin plot} 
\label{fig:UpperEnergyViolinPlot}
\end{wrapfigure}

The violin plot in Figure \ref{fig:UpperEnergyViolinPlot} summarizes the outlier energy across all 12 layers at convergence. For both the baseline MHA and Pre-RoPE MLA, the distribution is centered around $\approx 0.75$, with a long tail extending to $\sim 0.60$. This pattern suggests that approximately 75\% of the spectral energy remains concentrated in a few dominant directions---evidence of persistent rank compression. In contrast, the Decoupled MLA exhibits a significantly different trend: its distribution is both shifted downward and narrowed, with a median around $\approx 0.40$ and most of the mass concentrated between 0.20 and 0.55. This shift indicates that a substantial portion of the outlier energy has been redistributed into the MP bulk.

In summary, the violin plots confirm that {\em only the decoupled architecture effectively returns spike energy to the bulk}, thereby preserving a broader and more effective rank across layers.

{\bf Rotary budget and spectral stability}


\begin{wrapfigure}[13]{r}{0.48\textwidth  - .25\columnsep}
\centering
\vspace{-2\intextsep}
\includegraphics[width=.49\textwidth]{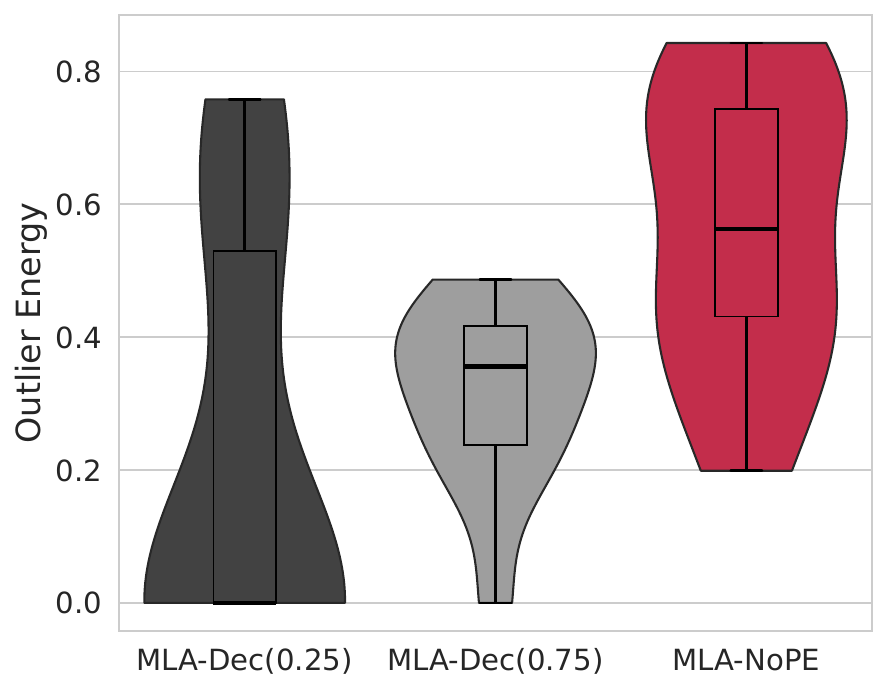} 
\vspace{-2.5em}
\caption{Upper energy  violin plot for various rotatory (RoPE) budget.} 
\label{fig:UpperEnergyViolinPlotNoPE}
\end{wrapfigure}

Figure \ref{fig:UpperEnergyViolinPlotNoPE} illustrates how reallocating head dimensions between content and RoPE affects the outlier-energy spectrum in Decoupled mode. The deviation from balanced RoPE allocation (50\% RoPE), Dec-0.25 and Dec-0.75, raises the spectral mass toward the outliers, signaling the reappearance of modest spikes. Nonetheless, the extreme case is \textbf{MLA-NoPE}, which lacks any positional encoding, and its spectrum is glued to the ceiling (median $\approx 0.80$, no tail), showing that $>$80\% of the energy collapses into a few dominant directions. Without a positional encoding, the model compresses both content and position into a narrow subspace, severely reducing representational diversity.

{\bf Perplexity Comparison}
Table \ref{tab:llama_ppl} summarizes the final perplexity values across MHA and MLA variants with different RoPE configurations. The balanced \textbf{Dec-0.50} matches MHA (26.86 vs. 26.89), while imbalanced settings (0.25/0.75) increase PPL by +0.15 to +0.20. The NoPE variant, dominated by spectral spikes, suffers a large degradation (+4.7 PPL to 31.54), highlighting the significance of rotatory embeddings. Recall that Decoupled MLA (Figure \ref{fig:MPGapStableRankHeatMaps}) eliminates the mid-layer MP-Gap spike and sustains more than 60\% Stable Rank across depth. Thus, positional encoding is essential for MLA, and a 50:50 content-to-position split is key to avoiding spectral bottlenecks that directly impair model quality.

\begin{table}[htbp]
\centering
\caption{Perplexity comparison across MHA and MLA variants with different RoPE configurations (LLaMA-130M). MLA without RoPE (NoPE) shows substantially degraded performance} \vspace{-0.5em}
\label{tab:llama_ppl}
\resizebox{0.99\textwidth}{!}{
\begin{tabular}{lcccccc}
\toprule
& MHA & MLA-Pre & MLA-Dec(0.25) & MLA-Dec(0.50) & MLA-Dec(0.75) & MLA-NoPE \\
\midrule
Eval PPL ($\downarrow$) & {\bf 26.89} & 27.72 & 27.02 &  \textcolor{blue}{\bf 26.86} & 27.07 & \textcolor{red}{\bf 31.54} \\
\bottomrule
\end{tabular}}
\end{table}

\section{Conclusion}

Our RMT analysis shows that sharing rotary embeddings across heads eliminates spectral spikes, maintains MP Gap at the noise level with outliers close to one, and preserves over 60 percent stable rank in MLA decoupled mode. In contrast, classical MHA and MLA Pre RoPE remain spike dominated and lose around 70 percent of spectral energy to a few dominant directions. 

{\bf Broader impact} Our work aims to bridges architectural efficiency with spectral interpretability. By combining memory-efficient attention mechanisms with RMT-based diagnostics, we uncover critical design insights for building future LLMs that are not only faster but also spectrally robust.

{\bf Limitations}
These findings are based on a 12 layer LLaMA-130M trained for 20K steps on 2.2B training tokens from C4 corpus. Heavier models, longer training schedules, or additional spectral metrics such as tail alpha may reveal new behaviors. Extending the logger to billion scale models and correlating spectral properties with downstream quality remain open directions for future work.

\bibliography{MyRef}

\begin{thebibliography}{23}
\providecommand{\natexlab}[1]{#1}
\providecommand{\url}[1]{\texttt{#1}}
\expandafter\ifx\csname urlstyle\endcsname\relax
  \providecommand{\doi}[1]{doi: #1}\else
  \providecommand{\doi}{doi: \begingroup \urlstyle{rm}\Url}\fi

\bibitem[Adlam et~al.(2022)Adlam, Levinson, and Pennington]{adlam2022random}
Ben Adlam, Jake~A Levinson, and Jeffrey Pennington.
\newblock A random matrix perspective on mixtures of nonlinearities in high
  dimensions.
\newblock In \emph{International Conference on Artificial Intelligence and
  Statistics}, 2022.

\bibitem[Bao et~al.(2024)Bao, Hataya, and Karakida]{bao2024self}
Han Bao, Ryuichiro Hataya, and Ryo Karakida.
\newblock Self-attention networks localize when {QK}-eigenspectrum
  concentrates.
\newblock In \emph{Proceedings of the 41st International Conference on Machine
  Learning (ICML)}, 2024.

\bibitem[Bouchard et~al.(2024)Bouchard, Mian, Tiomoko, Ginolhac, and
  Pascal]{bouchard2024random}
Florent Bouchard, Ammar Mian, Malik Tiomoko, Guillaume Ginolhac, and Frederic
  Pascal.
\newblock Random matrix theory improved fr\'echet mean of symmetric positive
  definite matrices.
\newblock In \emph{Forty-first International Conference on Machine Learning},
  2024.

\bibitem[Couillet et~al.(2016)Couillet, Wainrib, Ali, and
  Sevi]{couillet2016random}
Romain Couillet, Gilles Wainrib, Hafiz~Tiomoko Ali, and Harry Sevi.
\newblock A random matrix approach to echo-state neural networks.
\newblock In \emph{International Conference on Machine Learning}, 2016.

\bibitem[Dandi et~al.(2025)Dandi, Pesce, Cui, Krzakala, Lu, and
  Loureiro]{dandi2025a}
Yatin Dandi, Luca Pesce, Hugo Cui, Florent Krzakala, Yue Lu, and Bruno
  Loureiro.
\newblock A random matrix theory perspective on the spectrum of learned
  features and asymptotic generalization capabilities.
\newblock In \emph{The 28th International Conference on Artificial Intelligence
  and Statistics}, 2025.

\bibitem[Feofanov et~al.(2023)Feofanov, Tiomoko, and
  Virmaux]{feofanov2023random}
Vasilii Feofanov, Malik Tiomoko, and Aladin Virmaux.
\newblock Random matrix analysis to balance between supervised and unsupervised
  learning under the low density separation assumption.
\newblock In \emph{International Conference on Machine Learning}, 2023.

\bibitem[Firdoussi et~al.(2025)Firdoussi, Seddik, Hayou, ALAMI, Alzubaidi, and
  Hacid]{firdoussi2025maximizing}
Aymane~El Firdoussi, Mohamed El~Amine Seddik, Soufiane Hayou, Reda ALAMI, Ahmed
  Alzubaidi, and Hakim Hacid.
\newblock Maximizing the potential of synthetic data: Insights from random
  matrix theory.
\newblock In \emph{The Thirteenth International Conference on Learning
  Representations}, 2025.

\bibitem[Ilbert et~al.(2024)Ilbert, Tiomoko, Louart, Odonnat, Feofanov,
  Palpanas, and Redko]{ilbert2024analysing}
Romain Ilbert, Malik Tiomoko, Cosme Louart, Ambroise Odonnat, Vasilii Feofanov,
  Themis Palpanas, and Ievgen Redko.
\newblock Analysing multi-task regression via random matrix theory with
  application to time series forecasting.
\newblock \emph{Advances in Neural Information Processing Systems}, 2024.

\bibitem[Levi and Oz(2023)]{levi2023underlying}
Noam Levi and Yaron Oz.
\newblock The underlying scaling laws and universal statistical structure of
  complex datasets.
\newblock \emph{arXiv preprint arXiv:2306.14975}, 2023.

\bibitem[Li et~al.(2025)Li, Yin, and Liu]{li2025mixln}
Pengxiang Li, Lu~Yin, and Shiwei Liu.
\newblock Mix-{LN}: Unleashing the power of deeper layers by combining pre-{LN}
  and post-{LN}.
\newblock In \emph{The Thirteenth International Conference on Learning
  Representations (ICLR)}, 2025.

\bibitem[Liao and Couillet(2018)]{liao2018dynamics}
Zhenyu Liao and Romain Couillet.
\newblock The dynamics of learning: A random matrix approach.
\newblock In \emph{International Conference on Machine Learning}, 2018.

\bibitem[Liu et~al.(2024{\natexlab{a}})Liu, Feng, Wang, Wang, Liu, Zhao, Dengr,
  Ruan, Dai, Guo, et~al.]{liu2024deepseek}
Aixin Liu, Bei Feng, Bin Wang, Bingxuan Wang, Bo~Liu, Chenggang Zhao, Chengqi
  Dengr, Chong Ruan, Damai Dai, Daya Guo, et~al.
\newblock Deepseek-v2: A strong, economical, and efficient mixture-of-experts
  language model.
\newblock \emph{arXiv preprint arXiv:2405.04434}, 2024{\natexlab{a}}.

\bibitem[Liu et~al.(2024{\natexlab{b}})Liu, Feng, Xue, Wang, Wu, Lu, Zhao,
  Deng, Zhang, Ruan, et~al.]{liu2024deepseekV3}
Aixin Liu, Bei Feng, Bing Xue, Bingxuan Wang, Bochao Wu, Chengda Lu, Chenggang
  Zhao, Chengqi Deng, Chenyu Zhang, Chong Ruan, et~al.
\newblock Deepseek-v3 technical report.
\newblock \emph{arXiv preprint arXiv:2412.19437}, 2024{\natexlab{b}}.

\bibitem[Marchenko and Pastur(1967)]{marchenko1967distribution}
VA~Marchenko and Leonid~A Pastur.
\newblock Distribution of eigenvalues for some sets of random matrices.
\newblock \emph{Mat. Sb.(NS)}, 72\penalty0 (114):\penalty0 4, 1967.

\bibitem[Martin and Mahoney(2021)]{martin2021implicit}
Charles~H Martin and Michael~W Mahoney.
\newblock Implicit self-regularization in deep neural networks: Evidence from
  random matrix theory and implications for learning.
\newblock \emph{Journal of Machine Learning Research}, 2021.

\bibitem[Meng et~al.(2025)Meng, Yao, and Zhang]{meng2025transmla}
Fanxu Meng, Zengwei Yao, and Muhan Zhang.
\newblock Transmla: Multi-head latent attention is all you need.
\newblock \emph{arXiv preprint arXiv:2502.07864}, 2025.

\bibitem[Pennington and Bahri(2017)]{pennington2017geometry}
Jeffrey Pennington and Yasaman Bahri.
\newblock Geometry of neural network loss surfaces via random matrix theory.
\newblock In \emph{International conference on machine learning}, 2017.

\bibitem[Pennington and Worah(2017)]{pennington2017nonlinear}
Jeffrey Pennington and Pratik Worah.
\newblock Nonlinear random matrix theory for deep learning.
\newblock \emph{Advances in neural information processing systems}, 30, 2017.

\bibitem[Staats et~al.(2024)Staats, Thamm, and Rosenow]{staats2024locating}
Max Staats, Matthias Thamm, and Bernd Rosenow.
\newblock Locating information in large language models via random matrix
  theory.
\newblock \emph{arXiv preprint arXiv:2410.17770}, 2024.

\bibitem[Thamm et~al.(2024)Thamm, Staats, and Rosenow]{thamm2024random}
Matthias Thamm, Max Staats, and Bernd Rosenow.
\newblock Random matrix theory analysis of neural network weight matrices.
\newblock In \emph{High-dimensional Learning Dynamics 2024: The Emergence of
  Structure and Reasoning}, 2024.

\bibitem[Tiomoko et~al.(2019)Tiomoko, Couillet, Bouchard, and
  Ginolhac]{tiomoko2019random}
Malik Tiomoko, Romain Couillet, Florent Bouchard, and Guillaume Ginolhac.
\newblock Random matrix improved covariance estimation for a large class of
  metrics.
\newblock In \emph{International Conference on Machine Learning}, 2019.

\bibitem[Wei et~al.(2022)Wei, Hu, and Steinhardt]{wei2022more}
Alexander Wei, Wei Hu, and Jacob Steinhardt.
\newblock More than a toy: Random matrix models predict how real-world neural
  representations generalize.
\newblock In \emph{International conference on machine learning}, 2022.

\bibitem[Zhao et~al.(2025)Zhao, Deng, Ruan, Dai, Gao, Li, Zhang, Huang, Zhou,
  Ma, et~al.]{zhao2025insights}
Chenggang Zhao, Chengqi Deng, Chong Ruan, Damai Dai, Huazuo Gao, Jiashi Li,
  Liyue Zhang, Panpan Huang, Shangyan Zhou, Shirong Ma, et~al.
\newblock Insights into deepseek-v3: Scaling challenges and reflections on
  hardware for ai architectures.
\newblock \emph{arXiv preprint arXiv:2505.09343}, 2025.

\end{thebibliography}

\newpage
\clearpage
\appendix

\section{Attention Entropy Distribution in MHA and MLA}

\begin{figure} [htbp]
\centering
\subfigure[Attention entropy MHA \label{subfig:mha_attn_ent_hmap}]{\includegraphics[width=.32\textwidth]{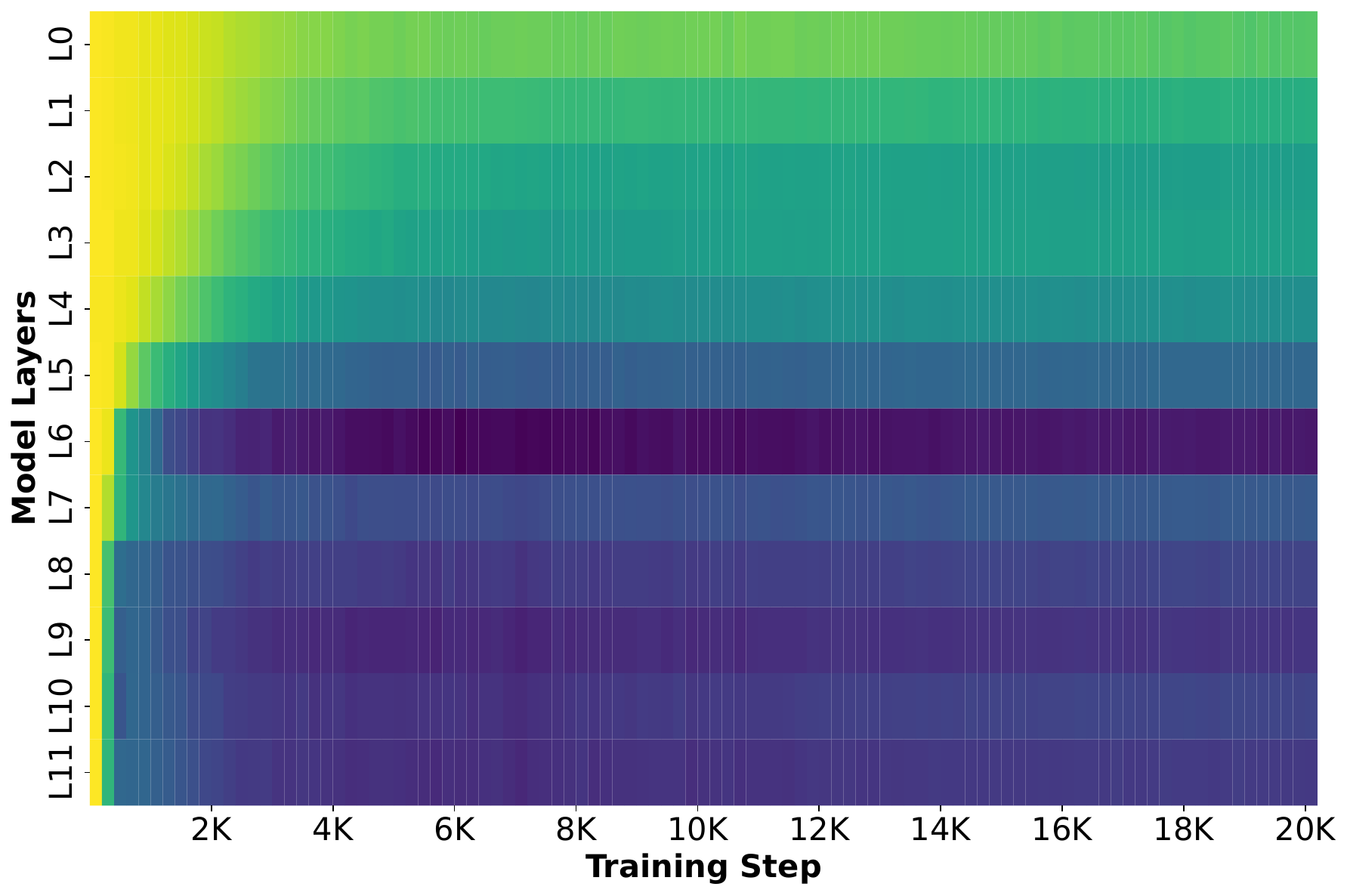}} 
\subfigure[MLA-Decoupled \label{subfig:mla_attn_ent_dec_r2}]{\includegraphics[width=.31\textwidth]{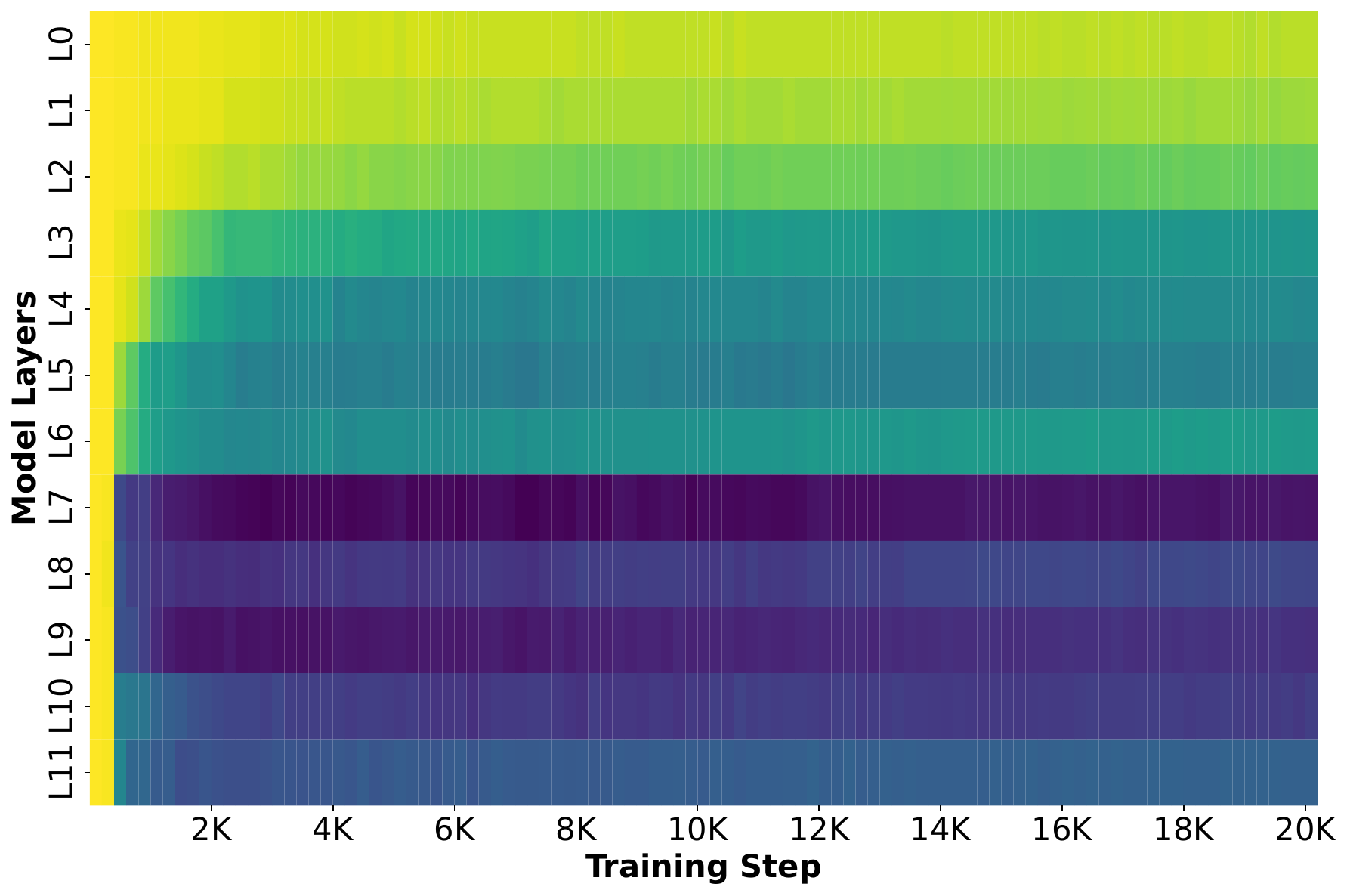}} 
\subfigure[MLA-PreRoPE \label{subfig:mla_attn_ent_pre_r2}]{\includegraphics[width=.31\textwidth]{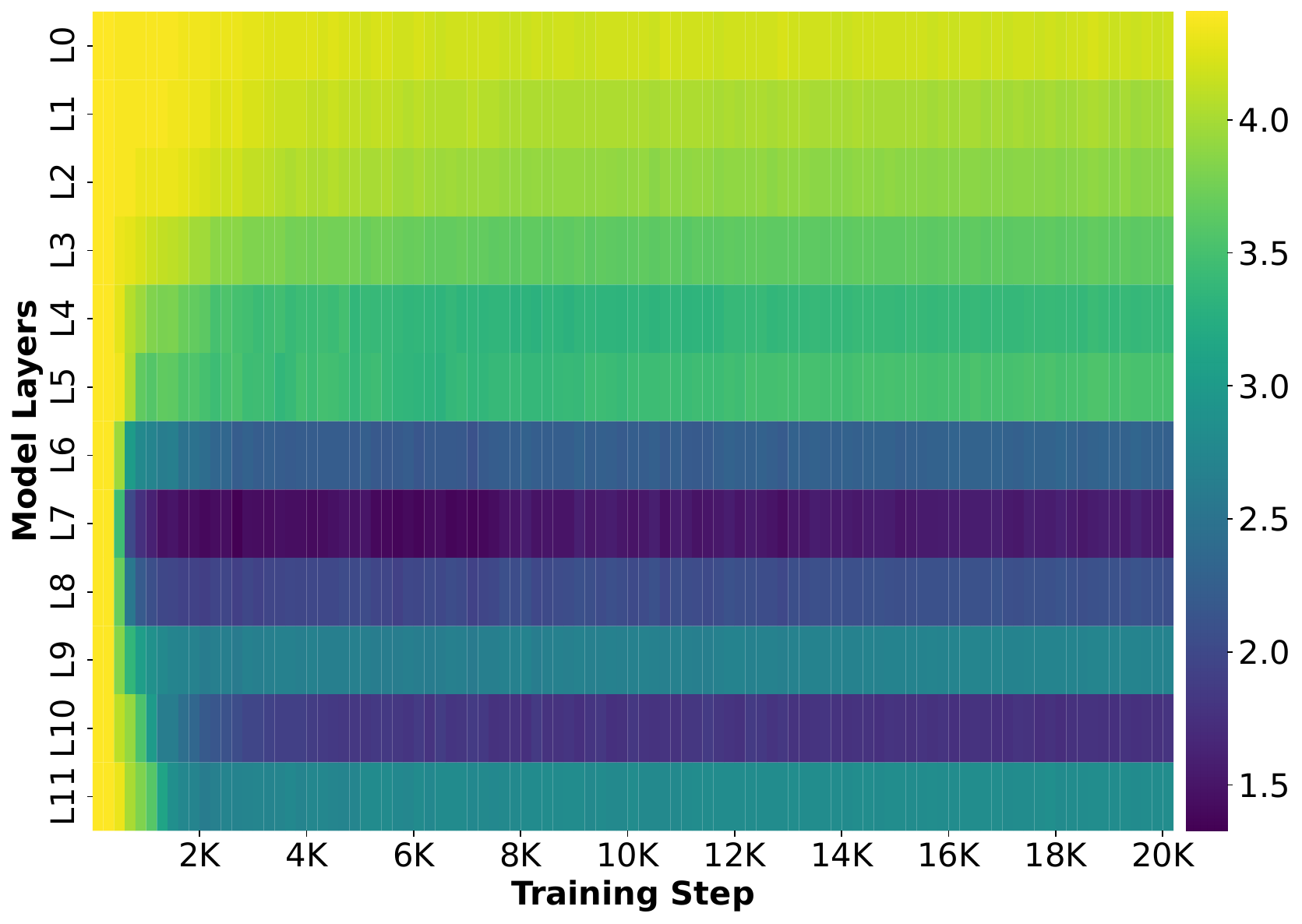}}  \\
\caption{Attention entropy patterns in classical MHA and MLA variants (decoupled and Pre-RoPE)} 
\label{fig:AttnEntropyHmaps}
\end{figure}

{\bf MLA improves information flow and stability across layers}
Our entropy analysis in Figure \ref{fig:AttnEntropyHmaps} highlights significant differences in information flow across three attention mechanisms. Vanilla MHA displays a clear bifurcation: early layers (L0-L3) quickly reach an entropic-overload state ($>$4 bits), while a deep entropy drop around layer 5 plunges below 1.5 bits and fails to fully recover. This rigid stratification suggests rich information flow at the network's start but significant starvation in the middle and deeper layers.

MLA-Decoupled softens these extremes, moderating both overload and starvation. MLA-PreRoPE further improves the entropy distribution: the middle-layer entropy dip nearly disappears, deeper layers recover rapidly within the first 5,000 steps, and the overall stack stabilizes twice as quickly as MHA. Thus, combining latent compression with pre-RoPE positional embeddings yields a more uniform and rapidly converging information flow, highlighting how nuanced architectural adjustments can significantly enhance transformer performance.

\section{Computational Cost}
All four diagnostics are computed from a single forward-pass SVD on the $W_{Q}W_{K}^\top$ weight matrix, imposing less than 1\% runtime overhead. For instance, an SVD on a $768 \times 768$ matrix costs only 3.6M FLOPs, negligible compared to a transformer forward pass. The method scales efficiently to larger models via subsampling of heads or layers, and leaves the backward graph untouched, ensuring that training throughput is virtually unaffected.

\end{document}